# Artificial Intelligence in Pediatric Echocardiography: Exploring Challenges, Opportunities, and Clinical Applications with Explainable AI and Federated Learning

Mohamed Yaseen Jabarulla[a†], Theodor Uden[b†], Thomas Jack[b], Philipp Beerbaum[b‡], Steffen Oeltze-Jafra[a‡*]

[a]*Peter L. Reichertz Institute for Medical Informatics of TU Braunschweig and Hannover Medical School, Hannover, Germany*
[b]*Department of Pediatric Cardiology and Intensive Care Medicine, Hannover Medical School, Hannover, Germany*

**Abstract**

Pediatric heart diseases present a broad spectrum of congenital and acquired diseases. More complex congenital malformations require a differentiated and multimodal decision-making process, usually including echocardiography as a central imaging method. Artificial intelligence (AI) offers considerable promise for clinicians by facilitating automated interpretation of pediatric echocardiography data. However, adapting AI technologies for pediatric echocardiography analysis has challenges such as limited public data availability, data privacy, and AI model transparency. Recently, researchers have focused on disruptive technologies, such as federated learning (FL) and explainable AI (XAI), to improve automatic diagnostic and decision support workflows. This study offers a comprehensive overview of the limitations and opportunities of AI in pediatric echocardiography, emphasizing the synergistic workflow and role of XAI and FL, identifying research gaps, and exploring potential future developments. Additionally, three relevant clinical use cases demonstrate the functionality of XAI and FL with a focus on (i) view recognition, (ii) disease classification, (iii) segmentation of cardiac structures, and (iv) quantitative assessment of cardiac function.

## 1. Introduction

Pediatric heart diseases include a broad range of structural abnormalities, cardiac muscle diseases, (pulmonary) vascular diseases and arrhythmias affecting infants and children. Especially cardiac malformations range from minor defects like a small atrial septal defect, to much more complex problems involving and affecting several parts of the pediatric heart, for example in a hypoplastic left heart syndrome [1]. Diagnosing, monitoring and decision-making in pediatric congenital heart diseases (CHD) necessitate a multidisciplinary and experienced team. Next to clinical and anamnestic parameters, various imaging modalities (CT, MRI, echocardiography) are particularly relevant to clarify the individual anatomy and hemodynamics [2]. Among these modalities, echocardiography plays a pivotal role in pediatric CHD diagnosis and the guidance of therapeutical decisions. Echocardiography utilizes ultrasound technology to generate real-time images of the heart, which does provide excellent information on heart structure and functional conditions [3]. Different from CT or MRI imaging, the machines can be directly brought to the patient allowing a fast and flexible diagnosis. Despite the numerous advantages, clinicians and sonographers encounter several challenges when utilizing this imaging technique. Firstly, obtaining accurate images that depict specific views in a standardized manner and display all relevant structures requires considerable expertise and manual adjustments for example of the correct angle of acquisition. Secondly, measurements are affected by a significant inter-observer bias since the acquisition and configuration of specific measurement points vary from one examiner to another [4]. Thirdly, in contrast to adults, the pediatric heart undergoes substantial age-dependent changes, demanding a thorough comprehension of its developmental trajectory. The strong longitudinal evolution adds a crucial layer of complexity to the interpretation of echocardiographic images. To overcome some of these limitations and improve diagnostic results, researchers focused on developing AI-based early diagnosis and intervention systems for echocardiography images [5,6].

†These authors contributed equally to this work and share first authorship
‡These authors contributed equally to this work and share last authorship
* Corresponding author
Email: †Jabarulla.Mohamed@mh-hannover.de, †Uden.Theodor@mh-hannover.de, *Oeltze-Jafra.Steffen@mh-hannover.de

AI might enable more efficient solutions to image interpretation than conventional imaging process workflows that heavily depend on expert experience and require a lot of workload to analyze the images. Furthermore, subtle image features that are not regularly detected to date might become recognizable. With the advancements of machine learning (ML) and deep learning (DL) algorithms [7,8], there have for example been attempts to implement fully-automated algorithms for echocardiography view classification, segmenting sections of the heart (ventricles, mitral valve, aorta, etc.) for functional assessment, and to perform quantitative assessments to measure the dimensions of the heart [9–11]. Automated measurements or assessments of heart structures help predict surgery outcomes by evaluating the measured values based on scoring systems [7,10]. Also in the field of fetal medicine, cardiac ultrasound is a relevant field of research [9,12]. Prior studies focused on applications like automatic detection of fetal cardiac position, automatic detection of the four-chamber (4CH) heart view, and segmenting of the different chambers for CHD. However, there are still significant challenges in adopting these algorithms and applications related to AI in echocardiography-based clinical decision support systems, such as the variability in image quality, the need for large annotated datasets, integration into existing clinical workflows, and ensuring the interpretability and transparency of AI decisions.

On the one hand, all algorithms depend on ground-truth image data also called 'training data set', which must be accurate and reproducible [13,14]. On the other hand, most DL models are trained using limited adult and pediatric echocardiography datasets [14]. To overcome the challenge of limited datasets and solve the problem of data privacy, researchers developed several *federated learning (FL)* paradigms allowing to collaboratively train AI models without sharing the original datasets [15,16].

The 'black box' nature of proposed AI algorithms makes it difficult for cardiologists to use these models for CHD diagnosis reliably [17] [18]. Here, the term 'black box' refers to AI systems whose internal processes and decision-making approaches are not transparent or easily interpretable, thereby complicating their integration into clinical practice. To take this circumstance into account, *explainable AI (XAI)* algorithms were developed to improve clinicians' understanding of and trust in AI-generated results [19,20]. Overall, there is a lack of comprehensive research specifically on utilizing FL and XAI in the context of pediatric echocardiography for increasing data availability and developing trustable diagnostic systems. This gap in the literature was the primary motivation for our research.

This paper aims to explore FL and XAI technology to mitigate the challenges associated with AI adoption for CHD decision support using echocardiography data:

1. We highlight the diverse spectrum of pediatric CHD and examine the challenges in echocardiography analysis for managing this complexity, aiming to provide a better understanding of the real-life conditions that technical approaches must address.
2. We review conventional AI-based solutions, their basic concepts, and recent trends in FL and XAI relevant to pediatric echocardiography.
3. We investigate the state-of-the-art research and challenges in adopting AI technologies for pediatric echocardiography datasets related to CHD.
4. We discuss the synergistic workflow and role of integrating FL and XAI in pediatric echocardiography providing researchers with insights to develop automated echocardiography-based decision support systems.

5. We discuss example clinical use cases based on FL and XAI focusing on view recognition, disease classification, segmentation of heart structures, and quantitative assessment of heart function for pediatric heart diseases.
6. We identify the limitations of the disruptive technologies FL and XAI and discuss future research directions that are anticipated to be of significant value in pediatric cardiology.

## 2. Background

*2.1. Echocardiography in Complex Pediatric Heart Conditions*

Echocardiography utilizes ultrasound technology to generate real-time images of the heart, providing detailed information about its structure, function, and blood flow [3]. It allows clinicians to visualize and assess cardiac abnormalities non-invasively and at the bedside, making it particularly suitable for pediatric patients. We highlight three examples of complex pediatric heart diseases that represent relevant clinical challenges to date that involve echocardiography. These example highlight a representative but still small spectrum of pediatric cardiology cases to understand the pivotal role of echocardiography. In the use case of *congenital aortic stenosis*, there is a narrowing of the heart valve between the left heart chamber and the aorta. If this narrowing is particularly severe, the heart valve must be dilated, e.g. utilizing a cardiac catheterization and an inflatable balloon, to allow a normal blood flow from the heart into the systemic circulation. In all cases, this intervention results in a regurgitation, which places a pathological burden on the left ventricle due to the blood flowing back. To counteract this situation, the heart valve is either replaced by an artificial one or tried to revise surgically. Determining the optimal timing for this operation is one of the major questions and regular echocardiographic examinations are necessary to measure flow velocities in the area of the heart valve to observe the function and size of the left ventricle.

In the case of *pulmonary hypertension* [21], there is a pathological change in the blood vessels of the lungs due to a wide variety of causes. Due to these changes, blood can only be pumped from the right heart through the lungs to a limited extent and with increased resistance. This situation puts particular stress on the right ventricle, as it is only designed to pump blood against lower resistances. In this case, the size and function of the right ventricle are in the focus of regular measurements. The special feature in this example is that the anatomy of the heart is nearly normal and any measurements must be taken with high precision and over time to identify even minor changes that might be relevant for therapeutic decisions.

*Hypoplastic left heart syndrome* is an example of a very complex and serious malformation of the heart [22,23]. Pathological changes may occur at various sites on the left side of the heart, such as a significantly reduced left ventricle size, valves that are too small, or an ascending aorta that is too narrow. In this clinical example, some cases cannot be assigned to the optimal surgical method because the pathological changes are only borderline and not full-blown. Deciding on the optimal surgical method is very challenging considering many relevant (echocardiographic) parameters that also change over time.

*2.2. Practical Challenges in a Pediatric Echocardiography Workflow*

Clinicians may encounter some limitations and challenges when using echocardiography to analyze these and other patient groups. These limitations and challenges include data acquisition, quantification, interpretation, and deriving prognostic information [9][22,24,25].

The **acquisition** of imaging from infants and less cooperative children is inherently more challenging. Pediatric patients often require specialized techniques and equipment adjustments to capture high-quality images. The ability to acquire and interpret these images can be hindered by the child's size and level of cooperation. **Standardizing** the acquired views is relevant for any objective evaluation. For this, the examiner must carefully adjust the settings of the echo machine and look for many different structures that must be displayed in the images correctly.

When it comes to **quantitative measurements**, assessing the size and function of different parts of the heart, like chambers or valves can be challenging. The pediatric heart undergoes significant changes as the child grows, which can affect echocardiographic measurements. Adjusting for these age-related changes requires expertise and experience in interpreting how normal development impacts echocardiographic findings. Especially when the heart is structurally abnormal, non-standard views are necessary to display the relevant structures. Therefore, especially in such cases, the probability is high that different examiners collect different measured values from the same patient and that objective comparability is limited (interobserver variability).

**Interpreting** the images and deriving a specific diagnosis might be a challenge as well, especially for examiners with limited experience. In some cases, diagnosis is easy to derive and only based on a single image, while in other cases certain structures must be acquired in different angles using different echocardiographic methods to fully get to the relevant information included. Furthermore, pediatric patients require longitudinal monitoring due to the rapid changes in heart size and function with age. Integrating longitudinal data to assess changes in heart size, function, and overall health is complex and demands careful evaluation of temporal variations in echocardiographic parameters. Finally, it is difficult to derive **prognoses or therapeutic consequences** from the available echocardiographic information because, especially in complex cardiac defects, the amount of information is large and weighting in this multifactorial context can be difficult.

## 2.3. Possible AI-based Solutions in Pediatric Echocardiography Workflow

Artificial intelligence (AI) has the potential to address the mentioned limitations for echocardiography-based decision support. Table 1 summarizes basic challenges and AI-based solutions for echocardiography. For instance, AI algorithms are trained to recognize patterns and identify anomalies in echocardiography images, reduce observer dependency, and improve the accuracy of diagnoses [9]. Additionally, AI algorithms are used to enhance echocardiography images, increasing the resolution, and allowing for the detection of small structural abnormalities. Moreover, AI algorithms can be trained on a large dataset of echocardiography images to provide a consistent and standardized interpretation of images in diagnosis [5,26].

**Table 1.** Summary of Echocardiography Challenges and AI-based Solutions

| Echocardiography Challenges | AI-based Solutions |
|---|---|
| Data acquisition and standardization | AI techniques can be trained to automatically acquire and standardize echocardiography data, reducing the need for manual input and improving the consistency and quality of the data |
| Quantification of cardiac structures | DL algorithms can automatically segment and measure cardiac structures, such as the left ventricle. This improves measurement accuracy and reproducibility |
| Interpretation of images | DL model trained to recognize patterns and identify anomalies in echocardiography images, reducing observer dependency and improving diagnosis accuracy |
| Prognosis and treatment planning | ML and DL algorithms can analyze large datasets of patient outcomes and predict prognosis and optimal treatment plans, helping clinicians make more informed decisions |

## 3. Innovative AI Approaches in Pediatric Echocardiography: From Basic Concepts to Clinical Applications

*3.1. Artificial Intelligence in Medical Imaging*

AI refers to machines that simulate human intelligence to think and act like humans. A computer program or algorithm is developed to perform tasks that would require human intelligence, such as recognizing patterns, making decisions, or resolving problems. AI has gained significant attention as a transformative technology that can impact various industries, including medicine [27–29]. AI can analyze large medical data to improve diagnosis, treatment, and patient outcomes by building an intelligent system using ML and DL techniques [30,31]. To fully grasp the role of AI in echocardiography for diagnosing various CHD, it is necessary to understand the conventional ML and DL approaches. ML is a subset of AI that focuses on creating algorithms enabling computers to learn from data and make informed predictions or decisions. Deep learning DL, a specialized branch of ML, utilizes multi-layered neural networks to learn intricate representations of data.

There are two main forms of ML: supervised and unsupervised [31,32]. Supervised learning involves the training of algorithms using a labeled dataset to make predictions about new data. Some popular techniques for supervised learning in medicine include Support Vector Machines (SVMs), k-Nearest Neighbors (k-NN), Random Forest, Naive Bayes, Gradient Boosting, and decision trees [33,34]. For instance, a ML algorithm is trained using a large dataset of labeled echocardiography images, each labeled as normal or abnormal [5]. The algorithm is then employed to analyze new echocardiography data and predict whether they are normal or abnormal images. This has the potential to greatly enhance the speed and accuracy of disease diagnosis in children.

Unsupervised learning [34], on the other hand, does not require labeled data. Instead, it searches for hidden structures in the data. For example, unsupervised learning is used in clinical applications to identify subgroups of patients with similar characteristics or to uncover hidden relationships in data. Unsupervised learning could be used for clustering patient data for more personalized treatment. Some popular techniques for unsupervised learning in medicine include hierarchical clustering, K-means clustering, and dimensionality reduction techniques such as t-Distributed Stochastic Neighbor Embedding (t-SNE) and Principal Component Analysis (PCA).

As aforementioned, DL is a form of ML that uses neural networks with multiple layers to learn from data. Neural networks, modeled after the human brain, enable DL algorithms to continually improve by analyzing data. It has been highly successful in a variety of applications, including medical image analysis and predicting patient outcomes [35,36]. One example of the application of DL in clinical data is the use of convolutional neural networks (CNNs) for image classification in radiology. A CNN can be trained on a large dataset of medical images to recognize, classify, and segment-specific diseases, such as tumors or lesions, based on their visual characteristics. The trained model is further used to analyze new images and provide a diagnosis with high accuracy. A few popular and successful CNNs algorithms that have been used for medical image analysis are U-Net, InceptionNet, ResNet, VGGNet, AlexNet, and LeNet [37]. On the other hand, TensorFlow, Keras, and PyTorch are open-source software libraries [38] that provide a platform for building and training DL models, including CNNs. These software libraries provide pre-trained models that can be fine-tuned for specific image diagnosis tasks and easily integrated into medical workflows. Another type of DL is recurrent neural networks (RNNs), which incorporate feedback by using the outputs from previous time steps as inputs for the current step. This mechanism enables the analysis of sequential data. In recent years, Vision Transformers (ViTs) [39] have emerged as a powerful alternative to CNNs for medical image analysis. By dividing images into patches and using transformer architectures, ViTs capture long-range dependencies more effectively, showing strong performance in tasks like classification, segmentation, and disease detection, especially with large datasets. Additionally, Foundation Models [40] have transformed AI by enabling large, pre-trained models to be adapted for specific medical tasks with minimal fine-tuning. Models such as CLIP (Contrastive Language-Image Pretraining), GPT (Generative Pre-trained Transformer), and Clinical-BERT (Bidirectional Encoder Representations from Transformers) excel at handling multimodal data, integrating both visual and textual information [41–43].

In short, ML and DL are related but distinct concepts. ML is the process of using algorithms to learn from data and make predictions, while DL is a specialized form of ML that uses deep neural networks to learn from data. With the advent of advanced techniques like Vision Transformers and Foundation Models, AI's capacity to transform industries like healthcare continues to grow, offering innovative solutions for tasks like medical image analysis, diagnosis, and personalized patient care.

*3.2. Recent Trends in Explainable AI and Federated Learning*

XAI and FL are gaining importance in AI for medical imaging, especially when combined [20,44,45]. XAI assists in opening 'black box' AI models by describing how certain recommendations are generated or results are obtained. This in turn enables the evaluation of these models to ensure they are fair, accurate, and compliant with regulations [18]. XAI is crucial for improving the transparency and interpretability of AI models. This is particularly important in healthcare [20] where it is important to understand the reasons behind model decisions. XAI techniques, such as saliency maps, feature importance, and rule-based explanations [45], allow users to gain insights into the model's decision-making process and identify biases, errors, or areas of improvement.

To understand the current trends in the implementation of XAI methods in medical imaging, a PubMed analysis was conducted by Borys et al. [46] based on manual categorization of all methods into visual and non-visual categories. Explainability methods based on visual explanations are frequently used to provide insights into a model's decision-making process [45]. These approaches rely on analyzing spatial information preserved through the convolutional layers of a model. This is done to identify which parts of an image are most salient to the resulting prediction. This leads to the generation of attribution maps, which highlight the most distinctive parts of an image that contribute to the final decision. These maps are typically represented as heatmaps, with different colors indicating positive and negative contributions to the activation of the target output [47]. Attribution maps can be generated using perturbation-based or backpropagation-based methods [19], both commonly used as post-hoc visualization approaches. Other back-propagation methods use network structures, including Activation maximization, Deconvolution, Class activation maps (CAM), and Grad-CAM, utilizing the activation function of the pooling layers or convolution layers to determine which input accounts for the particular output.

Non-visual approaches include auxiliary, textual, and case-based methods. Auxiliary measures provide insights in tabular or graphical formats, such as statistical indicators or feature importance, while textual methods offer semantic explanations of the model's predictions. Case-based explanations focus on identifying task-relevant concepts or influential samples [48]. Feature importance-based XAI methods like Local Interpretable Model-Agnostic Explanations (LIME) focus on generating interpretable explanations for the model's predictions by approximating the model locally with a simple interpretable model [49]. LIME approximates the model's behavior by building interpretable surrogate models around local instances and quantifies the impact of each feature on the model's output. Table 2 summarizes the various XAI methods used in medical image analysis, highlighting their purposes, and the limitations. We can observe from Table 2 that among existing XAI methods, visual explanations are the most frequently used in medical image analysis, likely due to their ease of interpretation.

**Table 2.** Summary of XAI Methods in Medical Image Analysis [46,49,56–58]

| XAI Method/Technique | Description | Limitations |
| --- | --- | --- |
| Saliency maps | Identifies the most saliency regions in an image that contribute to the model's decision. | May oversimplify complex decision boundaries. |
| Feature importance | Quantifies the importance of different features in the image for the model's prediction. | Limited to capturing linear relationships between features. |
| Rule-based explanations | Provides interpretable rules to explain the model's decision. | Prone to overfitting or memorization of training data. |
| Perturbation-based methods | Examines how model predictions change when specific parts of the image are altered or removed. | Can be computationally expensive for high-resolution images or complex models. Perturbations may introduce artificial artifacts or distortions. |
| Backpropagation-based methods | Visualizes the regions of an image that contribute to the model's decision. | Limited to interpreting models with differentiable architectures. |
| LIME | Offers local explanations for individual predictions, aiding in model interpretability. | Relies on random sampling, which may not capture the entire decision space. |
| Grad CAM | Highlights the regions of an image that are most influential in the model's decision. | Restricted to models with global average pooling layers. May exhibit sensitivity to input resolution or resizing. |
| SHapley Additive exPlanations (SHAP) | Offers unified explanations for predictions, considering interactions between features. | Can be computationally intensive, particularly for deep networks or large datasets. |
| Textual explanations | Provides explanations in natural language, making them easier to understand for non-technical users. | Requires additional natural language processing techniques. May not capture fine-grained details provided by visual explanations. |
| Auxiliary measures | Presents additional information, such as feature importance or statistical indicators, in a concise and interpretable manner. | May not capture complex relationships between features or image regions. Relies on predefined statistical indicators, which may not capture all relevant information. |
| Case-based explanations | Helps identify relevant cases or influential samples that affect the model's decision. | Requires a comprehensive and representative dataset for effective case-based reasoning. |

FL, on the other hand, addresses the issue of data privacy by allowing models to be trained on decentralized data sources without the need for data to be centralized. FL approach allows multiple organizations or entities to collaboratively train a ML or DL model without sharing their data. In a healthcare context, this is particularly important for preserving patient privacy and confidentiality. FL works by first training a base model on a central server using a small amount of data from each participating entity. The base model is then distributed to each entity, and each entity trains the model further on its own data. The updates are then sent back to the central server, where they are aggregated to improve the base model. This process is repeated until the model achieves the desired level of performance. Researchers' recent works have explored the applications of FL in medical images, and grapple with this problem. Adnan et al. [50] conducted a case study using histopathology images to demonstrate the effectiveness of differentially private FL in medical image analysis. They showed that distributed training with privacy can achieve comparable performance to conventional methods, making it a reliable framework for collaborative model development. Liu et al. [51] performed experiments on COVID-19 identification with chest X-ray images based on the basic FL framework. In this study [52], FL was performed using the NVIDIA Clara Train SDK [53]. As part of a FL system, they used differential-privacy techniques to protect patient data. Before sharing patient information with other clients, the data of each patient is encoded. A complex mathematical algorithm prevents the restoration of original datasets through reverse engineering. The FL model allows NVIDIA to achieve comparable segmentation performance without directly sharing institutional data.

In general, FL approaches are categorized into centralized and decentralized methods [54][55][29]. In centralized FL, a central server or aggregator coordinates the learning process by receiving local model updates from client devices and aggregating them into a global model. Decentralized FL, on the other hand, eliminates the need for a central server and relies on direct communication among clients to train a shared global model. These approaches enable collaborative learning while preserving data privacy and security, with centralized methods relying on a central server and decentralized methods utilizing peer-to-peer communication. Both approaches aim to enable collaborative learning while preserving data privacy and security, but they differ in terms of how the learning process is coordinated and how the models are shared and aggregated.

Additionally, FL can be categorized into three types: Horizontal Federated Learning (HFL), Vertical Federated Learning (VFL), and Federated Transfer Learning (FTL) [59]. HFL involves collaboration among institutions with the same feature space (same set of characteristics or attributes) but different samples (different individual data points). VFL involves institutions with different feature spaces (different sets of characteristics) but the same sample ID (same individual data points). FTL applies when institutions have different feature spaces and samples, leveraging transfer learning techniques to share knowledge.

Radiology departments benefit from FL since it facilitates collaboration without sharing private patient information. The domain adaptation challenge is one of the most prominent challenges when applying FL to medical images, specifically medical images. Typically, hospitals use a wide variety of imaging devices and methods, resulting in markedly different images from hospitals, which can lead to ML methods overfitting to non-semantic differences. Several FL techniques and methods have been adapted to address the unique challenges and requirements of the healthcare domain such as privacy, security, data heterogeneity, imbalances, and bias. Some of the techniques and algorithms used in FL for healthcare are summarized in Table 3. Apart from that, there are few commonly used frameworks in which some FL schemes have been implemented such as Federated AI Technology Enabler Framework (FATE), Open Federated Learning (OpenFL), TensorFlow Federated Framework (TFF), PySif, IBM Federated Learning (IBM FL), Flower, and Federated Learning Simulator (FLSim) [55].

**Table 3.** Techniques and Algorithms Adapted in Federated Learning for Healthcare [16,50,60–64]

| Technique/ Algorithms | Description | Advantages | Limitations |
|---|---|---|---|
| Federated Averaging | A widely used FL algorithm that aggregates local model updates. | Preserves data privacy and security. Allows scalability with a large number of clients. | Federated Averaging |
| Federated Stochastic Gradient Descent | Updates the global model using local stochastic gradient descent on each participating device. | Reduces communication overhead. | Sensitive to device heterogeneity. |
| Differential Privacy | Incorporates differential privacy to protect sensitive data during model aggregation. | Provides privacy guarantees against inference attacks. | Increased communication and computational overhead. |
| Federated Transfer Learning | Transfers knowledge from pre-trained models to improve the learning process on local data. | Efficient utilization of pre-trained models. | Challenging to handle domain shifts and variations in local data. |
| Secure Multi-Party Computation | Perform computations on encrypted data. Clients encrypt their local models and share encrypted model updates to achieve | Provides strong data privacy and security guarantees | Computationally expensive due to cryptographic |

| | secure model aggregation without compromising sensitive data. | during model aggregation. | operations. |
|---|---|---|---|
| Federated Adaptation | Adapts the global model to local datasets by incorporating local model updates selectively. | Allows adaptation to diverse data distributions. | Potential bias in local model updates. |
| Split Learning | Separates the model into two parts: a client-side encoder and a central server-side decoder. Only encoded features are shared, ensuring privacy. | Minimizes data transmission. Protects sensitive information. | Requires high-bandwidth communication between clients and server. Potential loss of model performance due to partial sharing. |
| Federated Meta-Learning | Trains local meta-learners on individual clients for faster learning and adaptation across multiple tasks or domains. | Enables quick adaptation and learning on new clients' data. | Complexity in designing meta-learning algorithms. |
| Federated Reinforcement Learning | Extends FL to reinforcement learning tasks, where Agents interact with the environment, learn optimal policies, and exchange experiences or policies with a central server to improve the global policy. | Allows collaborative learning in reinforcement learning scenarios. | Communication overhead in exchanging experiences or policies. |

*3.3. Artificial Intelligence-based Interpretation of Pediatric Echocardiography*

AI has been increasingly explored for its potential to improve the diagnosis of CHD in pediatric patients using echocardiography images. Despite the success of AI models in diagnosing CHD in adult patients using echocardiography [11,65], the application of AI in the pediatric population presents unique challenges due to differences in anatomy, physiology, and image quality. As a result, existing adult AI models [66] are not directly applicable to pediatric echocardiography images for CHD diagnosis [67][68].

To address these challenges, researchers have developed novel approaches that take into account the specific characteristics of pediatric echocardiography images. For example, some studies have focused on developing AI algorithms that can handle the variability in echocardiography image quality due to differences in patient size and age [69]. Other studies have explored the use of transfer learning techniques to adapt existing adult AI models (EchoNet-Dynamic) to pediatric echocardiography images [70,71]. ViTs are a newer, potentially powerful approach to interpreting echocardiography videos, as long as they can effectively represent broader contextual information [72,73]. Foundation models have currently found their way into echocardiography interpretation and deliver impressive results even without any task-specific training [74]. Table 4 summarizes a few studies to investigate the effectiveness of AI in diagnosing CHD in pediatric patients through echocardiography images. These studies have shown promising results, demonstrating that AI algorithms significantly enhance the accuracy of CHD diagnosis and efficiently support clinicians in decision-making. Furthermore, AI has the potential to overcome some of the challenges faced by clinicians, such as observer dependency, limited imaging resolution, and interobserver variability.

**Table 4.** Summary of the application of AI in Pediatric echocardiography for CHD

| Author/Year | Article Type | AI model | Study population/ Number of Echocardiography dataset | Use cases of AI | Limitations |
|---|---|---|---|---|---|
| Chen. et al. [75] / 2017 | Original research | RNN | 1231 fetus videos | Identification of different standard views from US videos | Generic framework and not tested on CHD patients. |
| Dong et al [76] / 2022 | Original research | CNN | 2032 4CH and 5000 non-4CH images | Automatic quality control of fetal US cardiac 4CH plane | Only focused on 4CH views and anatomical structures are not considered. |
| Arnaout et al. [77] / 2021 | Original research | Ensemble of neural networks trained using DL approach | 107,823 images obtained from 1326 US studies | To identify specific cardiac views and distinguish between normal hearts and complex CHD. | Not specific to CHD subtype and less details regarding DL model. |
| Truong et al. [78] / 2022 | Original research | Random forest (RF) algorithm | Tabular data formulated from 3910 examinations | To assess the presence or absence of CHD | Not used echo images or videos. Not specific to CHD subtype. |
| Arnaout et al. [79] / 2018 | Archive article | CNN | 493 normal hearts, 87 tetralogy of Fallot (TOF) and 105 HLHS images | To identify the five crucial views, measure and segment cardiac structures, and distinguish between a normal heart, HLHS, and TOF | Limited dataset, algorithm not publicly available for recreating the results, Clinically untested. |
| Siti Nurmaini et al. [80] / 2021 | Original research | Mask-RCNN | 1149 fetal heart images | To predict 24 objects, including four fetal heart standard views, 17 heart chamber objects per view, and three CHD cases. | Clinically untested, algorithm not publicly available for recreating the results and model transparency issue. |
| Dozen et al. [81] / 2020 | Original research | U-net with modified input data and VGG network | 615 frames annotated from 421 normal fetal cardiac ultrasound videos | To segment ventricular septum (VS) | Limited dataset, Not specified the impact of AI model in identifying VS defect. |
| Anda et al. [82]/ 2023 | Original research | CNN | 300 video files used to extract 6000 images (around 1000 per class) | To assist human experts in identifying the 4CH view, the left ventricle outflow tract view, the RV outflow tract view, and the three-vessel trachea view | Only used normal heart dataset. The accuracy of this AI model for detecting CHD cannot be evaluated from these data |
| Addison et al. [83]/2022 | Original research | CNN | 27948 images used from 100 patients | To autonomously perform view classification on pediatric echocardiographic images | Not specific to any CHD. |

| Zuercher et al. [71] /2022 | Original research | Transfer learning | Hemodynamically insignificant anomalies or normal cardiac anatomy data obtained from 321 patients | Identifying patients with reduced left ventricular ejection fraction (LVEF) and monitoring LVEF responses to treatment | The AI model only tested on 4CH views. |
|---|---|---|---|---|---|
| Dozen et al. [81] / 2020 | Original research | U-net with modified input data and VGG network | 615 frames annotated from 421 normal fetal cardiac ultrasound videos | To segment ventricular septum (VS) | Limited dataset, Not specified the impact of AI model in identifying VS defect. |

These AI studies generally follow a standardized approach (Fig. 1) consisting of four stages, namely data collection, preprocessing, model development, and model testing. Through this approach, AI is used to extract relevant features from echocardiography data to recognize standard views, segment cardiac structures, evaluate cardiac function, and analyze prognosis in conjunction with multiple parameters.

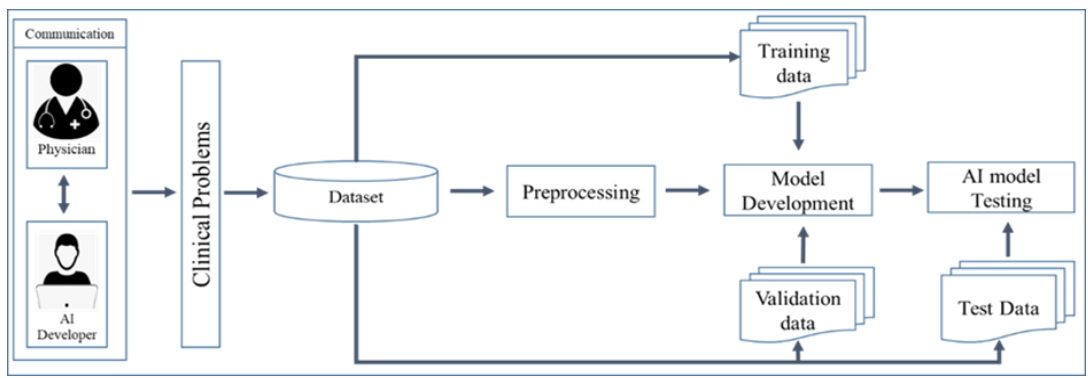

**Fig. 1.** AI Model Development in Echocardiography: A Five-Step Workflow

The Fig.1 illustrates a five-step workflow for developing AI models in echocardiography, which consists of five main steps: (1) Communication between physicians and AI can also be included as an essential step in the process, which can help to refine the clinical problem statement and guide the development of the AI models.; (2) collecting relevant clinical data; (3) pre-processing the data to meet task-specific requirements such as labelling or segmentation, and dividing them into training, validation, and testing datasets sets; (4) selecting appropriate AI algorithms for model development based on the task type, training the models on the training dataset, and validating their performance on the validation dataset; and (5) testing the reliability and generalization of the models on internal and external testing datasets.

Despite these efforts, there is still much work to be done to develop robust and accurate AI models for diagnosing CHD in pediatric patients using echocardiography. Table 2 also depicts that only a limited number of CHD conditions have been considered. There is still a research gap and opportunities in taking advantage of this technology for fully automated diagnoses of vast pediatric CHD. Furthermore, the incorporation of XAI and FL techniques, which are currently underutilized in pediatric echocardiography for diagnosis, presents a crucial research gap that needs to be addressed. In addition, there are data availability, AI model transparency, and ethical challenges that must be taken into account to ensure the safe and effective use of AI in clinical practice.

*3.4. Challenges in Utilizing AI for Pediatric Echocardiography*

This section highlights the key challenges and possible solutions. The significant challenges must be addressed to make AI diagnosis reliable and safe for pediatric patients.

    (a)  **Lack of standardization of echocardiography protocols across different medical centers.**

Standardizing echocardiography across different medical centers is challenging because each center may use different imaging techniques [84]. For example, one clinic might perform an echocardiogram from an upside-down angle, while another uses a standard top-to-bottom view. This variability can lead to inconsistent image quality and interpretation. Instead of trying to standardize all imaging procedures across clinics, a more effective approach is to select the highest-quality images from existing video recordings. For instance, if a center records a 30-second video of the heart, researchers can choose the clearest frames that show the heart's structures, regardless of the angle. This could include selecting images from both the upside-down and top-to-bottom views that best represent the heart's anatomy. By focusing on these high-quality images, researchers can create a more reliable dataset for training AI models. This ensures that the AI learns from the clearest and most consistent examples of how a healthy heart looks, no matter the original viewing angle. Ultimately, this method can enhance the accuracy of AI diagnostics for pediatric patients, leading to safer and more reliable outcomes.

    (b)  **Lack of large, high-quality, annotated datasets.**

The availability of large datasets is essential for AI model training and validation. However, in the case of pediatric CHD, there is a shortage of high-quality, annotated datasets due to the difficulty and time-consuming nature of manual annotation. FL addresses the lack of large, annotated datasets by enabling model training on decentralized data while preserving patient privacy. In addition, Generative Adversarial Networks (GANs) [85] can be employed to generate synthetic data that mimics the characteristics of the desired echocardiography dataset. This synthetic data, combined with the limited annotated data, can be utilized to create a larger and more diverse dataset for training DL models. Another solution involves the use of unsupervised techniques, which do not require labeled data, allowing models to learn from the inherent structure of the data. Additionally, foundation models, such as BERT, GPT-3, and Vision Transformers [39,40], can be leveraged as powerful pre-trained architectures. These models can enhance performance by transferring knowledge from vast datasets, facilitating more effective training in scenarios with limited labeled data.

    (c)  **Interpretability and transparency of AI models.**

The 'black box' nature of AI models makes it difficult to understand how the model arrived at its conclusions. This is a major concern for pediatric patients, as misdiagnosis can lead to devastating consequences. XAI techniques provide interpretability and transparency, enabling clinicians to understand the reasoning behind AI model diagnoses and reducing the risk of misdiagnosis [86].

    (d)  **Generalization of AI models across different medical centers.**

AI models developed in one medical center may not generalize to other centers due to variations in imaging protocols and patient populations. This challenge can be addressed by a federated transfer learning approach [64], which allows for the generalization of AI models across different medical centers by transferring knowledge from one model to another and accommodating variations in imaging protocols and patient populations.

Research on integrating XAI with FL models has received much less attention, but interest in this area is rapidly gaining traction. FL and XAI integration in medical image applications offers several benefits, such as improved trust and interpretability of the models, which are crucial for clinical decision-making. The specific challenges of integrating FL with XAI, as opposed to standard DL with XAI, include the complexity of ensuring explainability across distributed and heterogeneous data sources, maintaining data privacy while generating interpretable explanations, and addressing the computational overhead associated with XAI techniques in a federated setting.

One of the primary challenges is the heterogeneity of data across different institutions, which can affect the consistency and reliability of the explanations generated by XAI methods. Ensuring that the explanations remain robust and meaningful across varied datasets is a significant hurdle. Additionally, preserving patient privacy while providing detailed and accurate explanations is crucial, as XAI methods often require access to comprehensive data to generate insightful interpretations.

To tackle these challenges, one promising approach is to implement federated model distillation, which allows local models to aggregate knowledge into a global model. This method enhances both performance and interpretability while protecting data privacy. Developing frameworks that prioritize explainability from the outset can also address the complexities in distributed data; for instance, utilizing techniques such as attention mechanisms and model-agnostic methods can improve interpretability without compromising privacy.

Furthermore, leveraging privacy-preserving techniques like differential privacy and secure multiparty computation can ensure patient privacy while still enabling XAI methods to provide meaningful insights. Establishing standardized data formats across institutions can enhance data consistency, improving the integration of XAI methods and leading to more reliable explanations. Creating iterative feedback loops where clinicians provide input on generated explanations can also enhance the relevance and trustworthiness of the models, ensuring they align closely with clinical needs. Additionally, developing adaptive algorithms that can learn from the specific characteristics of different datasets will enhance the robustness and relevance of the generated explanations. By implementing these strategies, the integration of FL and XAI in medical imaging can overcome significant barriers, ultimately leading to improved diagnostic accuracy, enhanced patient safety, and greater trust in AI systems among healthcare professionals.

*3.5. Potential of Integrated Explainable AI and Federated Learning in Decision Support*

The integration of XAI with FL represents a powerful approach to enhancing decision support systems across various medical modalities. This integration is especially significant in pediatric echocardiography due to the unique challenges and requirements of this field. Pediatric echocardiography involves high variability in data due to diverse congenital heart defects, necessitating robust and interpretable models for accurate diagnosis.

FL allows multiple medical institutes to collaboratively train AI models without sharing patient data, thus safeguarding data privacy. Each institute trains an AI model using locally acquired echocardiography data and shares the model weights with a central federated server. This server aggregates the models to create a global model, ensuring that the benefits of diverse datasets are realized without compromising data privacy. XAI ensures that these models remain interpretable, promoting ethical AI practices by ensuring accountability, fairness, and reduced biases.

Fig. 2 illustrates the workflow and role of integrated XAI and FL. The XAI with FL framework is organized into four distinct modules: Communication, Central Authority, Local Servers, and User Interface. The communication module involves collaboration between participants, such as physicians, clinicians, AI developers, and researchers. They work together to address the clinical problems associated with the decision-making process for pediatric CHD. Further, they focus on discussing AI tasks, ethical considerations, and creating echocardiography data preprocessing guidelines to ensure standardized data preparation. This includes anonymization, data cleaning, and format conversion. The collaborative approach ensures that the AI and FL process aligns with medical ethics and regulatory standards.

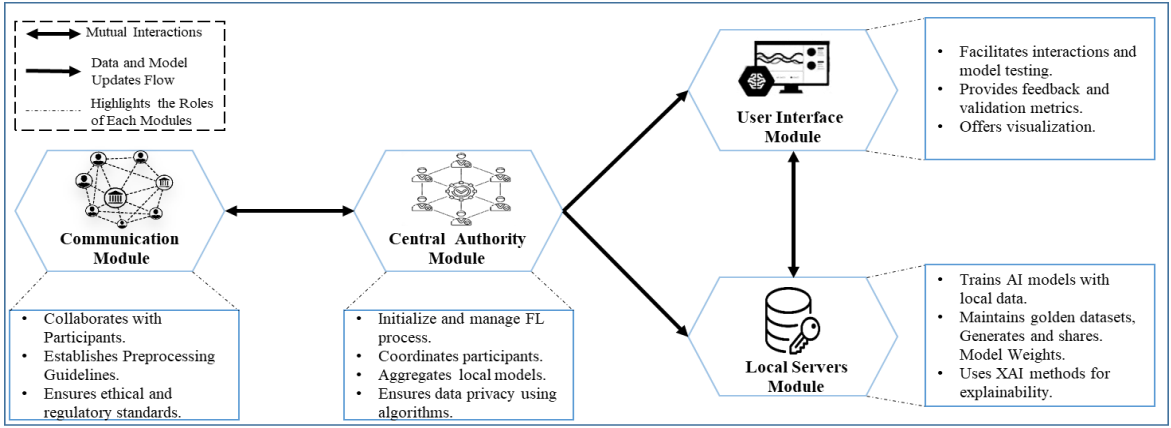

**Fig. 2.** Illustrates the four key modules and its role in the integration process: Communication Module, Central Authority Module, Local Servers Module, and User Interface Module.

The central authority module is responsible for initializing and overseeing the FL process. It acts as the central server that coordinates the communication and collaboration among the local servers. The central authority sets up the FL infrastructure, defines the FL protocol, and manages the overall process. This module incorporates the FL approach, where a central federated server coordinates the collaboration between multiple local servers (participants). Depending on the AI task, each local server trains a local model on its own echocardiography data. This is done without sharing the raw data with the central server or with other participants. After each round of local model training, the federated server collects the updated models from the local servers and performs model aggregation to create a global model. This module ensures that the knowledge from different local models is aggregated to enhance the overall performance. In addition, this module establishes security measures to protect data privacy, ensuring that data transmission and aggregation are performed securely. For example, researchers have proposed using secure multiparty computation (SMC) techniques to protect data privacy during FL [63]. SMC enables each local model can encrypt its data and send it to a trusted third party. The third party can then combine the encrypted data from all the local models and decrypt it to obtain an aggregate model. This process ensures that the data remains private and secure while still allowing the researchers to gain insights from the aggregated models. Wei et al. [62] have proposed the use of differential privacy to protect data privacy during the aggregation process so that data from other researchers cannot be accessed. Differential privacy adds noise to the collected data, making it difficult for outsiders to gain access to the data. This ensures that data privacy is maintained while still allowing researchers to access the aggregated data.

The local server module consists of N (multiple) local servers, each representing participating stakeholders [29] in the collaborative learning process. Each local server possesses its own echocardiography and clinical dataset, known as the golden dataset. A golden dataset is a high-quality and carefully curated dataset that serves as a reliable reference or benchmark for a specific task. In the context of XAI with FL for CHD decision support, it consists of annotated echocardiography data or relevant medical data representing a wide range of CHD cases. The golden dataset is used to train and evaluate AI models, serving as a reference for comparing the performance of local models created by participants in the FL process. Its creation involves expert knowledge and rigorous validation procedures to ensure accuracy and reliability. The participants of the local server module use their own dataset to train and validate the XAI models for specific tasks, such as segmentation, quantification, or classification. Depending on the allocated AI task and datasets, the XAI model generates explanations using methods such as saliency maps, Grad-CAM, or LIME [49]. For example, consider a segmentation task where we aim to automatically identify the LV for HLHS patients, where the LV size may range from normal to entirely absent. The LV is a crucial anatomical structure in the heart, and accurate segmentation is important for various clinical applications. Interpretable AI methods for this task are essential to understand how the model identifies and delineates the LV region. The interpretation of the model's segmentation results and explanations can be done using LIME. LIME generates explanations in the form of superpixel-based masks, highlighting the important regions that contribute to the segmentation outcome. It provides insights into the model's behavior and helps identify any biases or errors in the segmentation process. The local explanations generated by each participant were examined to gain insights into the important features or regions considered by the models for LV segmentation. This step helps ensure that the individual models are accurately segmenting the LV regions for example in the HLHS echocardiography images. The participants communicate with the central authority server by sharing trained model weights or sending model updates after each round of local training. The central server aggregates these updates and performs global validation to evaluate the model's performance on a diverse validation dataset.

The user interface model acts as an API that facilitates interactions between the participants involved in central authority and the local servers. The user interface is used to test the model on new data, view the results of each XAI model, and monitor the progress of the collaborative learning. The communication module able to provide feedback to the AI developers, which used to improve the model and refine the data preprocessing stage. It also provides a platform for performing global model validation and interpretation. Physicians and AI developers can access the user interface to evaluate the performance of the global model, validate its accuracy on unseen data, and interpret the model's decisions. The user interface may include visualizations, metrics, and interpretability techniques to aid in the analysis and diagnosis of echocardiography datasets for CHD. In the end, this module aims to provide flexibility and customization options in the user interface, enabling users to adapt it to their specific needs and preferences. For example, the interface can display saliency maps generated by the explainable AI component and provide a visual representation of the model's decision-making process [87].

*3.6. Application of Artificial Intelligence Solutions in Example Clinical Use Cases*

In routine echocardiography, a vast amount of potentially diagnostic information often remains underutilized due to the challenges of interpreting such a large volume of data within a limited timeframe [65]. AI algorithms can effectively unlock the value of these hidden findings and analyze the information at a faster pace to provide decision-making support to clinicians. Consequently, by integrating FL and XAI techniques, the framework enhances the analysis of echocardiographic images and provides valuable insights into various aspects of CHD diagnosis and management. In general, the framework aids in the diagnosis and analysis of complex congenital heart lesions. View recognition algorithms ensure the accurate identification of different echocardiographic views. Segmentation models can accurately segment and quantify specific cardiac structures, such as the ventricles, great vessels, or valves. Disease classification algorithms can differentiate between various complex lesions. The heart defects described as examples in section 2.1 illustrate to a particular extent the importance of the framework techniques described.

**Pulmonary Hypertension** [21]**:** The integrated workflow aids in evaluating pulmonary hypertension in pediatric patients with pediatric heart diseases. View recognition techniques identify the four-chamber view and the short-axis view on the ventricular level being the most relevant views for this condition. This ensures that the subsequent analysis is performed on appropriate images, enhancing the accuracy of the evaluation. Segmentation algorithms can delineate especially the size and function of the right ventricle. The right ventricular pressure can be additionally estimated by measuring the flow velocity of a blood flow jet at the tricuspid valve. Disease classification models can categorize pulmonary hypertension based on its severity. The gathered information assists in diagnosis, monitoring of the disease progression, and guiding treatment decisions, including the choice of medications.

**Borderline Hypoplastic Left Heart Syndrome (bHLHS)** [23]**:** The combination of XAI with FL is uniquely suited for evaluating cases of borderline HLHS in pediatric patients. This condition can impact multiple segments of the left side of the heart and the proximal aorta, necessitating a diverse range of less standardized cardiac views for accurate classification. Segmentation models play a pivotal role in delineating structures like the left ventricle, aortic valve, and aorta to assess their sizes. Additionally, disease classification algorithms might be valuable in "borderline" cases, aiding in determining whether the patient's smaller left ventricle has an at least still adequate size in order to be able to fulfill its function. Decisions regarding the appropriate course of action are subsequently based on this assessment. Given the extreme rarity of this cardiac condition, the concept of federated learning becomes significantly important.

**Aortic stenosis** [88]**:** The described framework provides the examiner with relevant information throughout the course of the disease. View recognition focuses in particular on the left ventricle. During the segmentation of the chamber, an assessment of function is also performed. In addition, the narrowed area of the aortic valve is precisely measured and the flow velocities measured in the Doppler procedure in the area of the valve are registered. A disease classification allows a categorization according to the severity and - with regular assessment during the course of the disease - indications as to when an optimal time for surgical valve replacement might be.

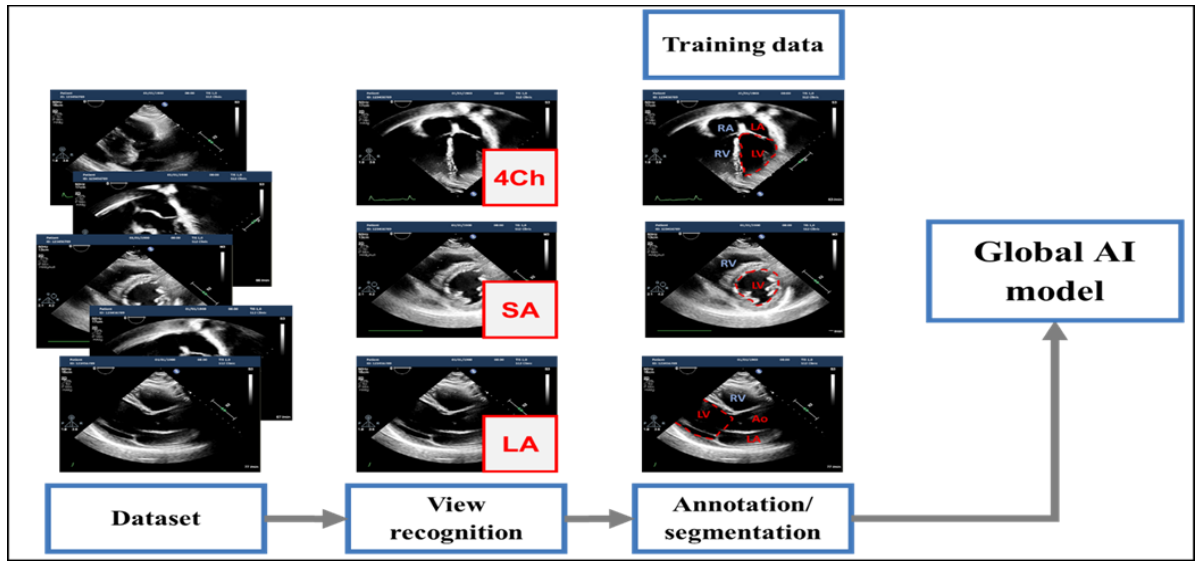

**Fig. 3.** Data preparation for segmentation tasks consists of view recognition and annotation/segmentation of relevant parts of the patient´s heart. Depending on the clinical use case, different regions of the heart might be relevant to be depicted and segmented.

Fig. 3 shows relevant steps necessary to prepare the echocardiographic data before applying to the global AI model. XAI techniques provide clinicians with visual explanations, highlighting areas of the echocardiogram that influence the diagnostic decision. This interpretability allows clinicians to understand the basis of the model's predictions, improving trust and facilitating clinical decision-making. The ability to interpret the model's reasoning enhances confidence in the diagnosis and helps guide appropriate treatment strategies. Additionally, the FL approach allows multiple healthcare institutions to collaborate and contribute their data, ensuring a diverse and representative dataset for training robust models.

## 4. Discussion

*4.1. Challenges in Including XAI together with FL and Possible Solutions*

Integrating XAI with FL offers promising opportunities in these clinical use cases, but there are a few challenges and limitations to consider. These include ensuring data privacy and security, addressing biases and interpretability issues in AI models, and validating the performance of the combined framework in diverse patient populations. Continued research, collaboration between clinicians and AI experts, and adherence to ethical guidelines are crucial to overcome these challenges and fully harness the potential of FL with XAI in pediatric echocardiography. In this section, we discuss some challenges that need to be addressed for developing an automated decision support system and leveraging the maximum benefit of the combined approach.

The first challenge is the security concerns while sharing and exchanging model updates between local servers and central servers. To address this, secure communication protocols, encryption techniques, and differential privacy can be implemented [50]. These techniques help to protect the confidentiality and integrity of the exchanged model updates without compromising the privacy of the underlying data. Additionally, strict access controls and data anonymization practices should be in place to ensure privacy. The second issue is the communication overhead. The communication between the central authority and local servers can introduce communication overhead, especially in scenarios where the number of participants is large or the network connectivity is limited. Hence, it is necessary to optimize communication protocols, compressing model updates, and minimizing data transmission can help mitigate this limitation [89]. The third challenge is heterogeneous data distribution and model generalization. Each participant in the local server may have imbalanced or heterogeneous datasets, resulting in variations in model performance across participants. This also affects the global model trained on the federated server leading to the model not generalizing well to unseen data. To address this, techniques such as data augmentation, data balancing, model aggregation, ensemble methods, or federated transfer learning can be employed to ensure fair model training across participants. Reasonable interpretability and explainability are the fourth issue. Although XAI algorithms provide interpretability, there may be challenges in interpreting the combined FL model due to its distributed nature. Aggregating explanations from local models to create a global interpretable model help address this limitation. In the end, to overcome these aforementioned challenges, it is crucial to establish collaboration between AI researchers, healthcare institutes, and regulatory bodies to define and implement best practices, guidelines, and standards. Continual monitoring, evaluation, and improvement of the framework are essential to ensure its effectiveness, privacy preservation, and ethical compliance in the context of automated decision support

*4.2. Future Work*

In future developments of decision support systems for pediatric echocardiography, several key aspects will be addressed. Integrating advanced AI and FL algorithms includes exploring state-of-the-art DL architectures, leveraging transfer learning approaches, and incorporating multimodal data sources for more comprehensive analysis. FL and explainable AI serve as an overarching concept allowing an integration into different frameworks. Building upon our preliminary work [90], the research can be extended by incorporating these concepts. FL and XAI will enhance tasks like image segmentation, disease state identification, and parameter extraction. Finally, we intend to validate and refine the integrated workflow using multimodal clinical datasets, demonstrate its potential in real-time clinical settings, and contribute to advancing the field of pediatric echocardiography and CHD diagnosis.

The differences in congenital heart anomalies, patient age, and clinical management underscore the need for tailored solutions. While utilizing XAI with FL may find applications in adult CHD or other cardiac conditions, its primary intention is to address the specific challenges and intricacies associated with pediatric CHD. We believe that a focused approach is essential to provide the pediatric cardiology community with an interpretable diagnostic tool, while future research endeavors can explore adaptations for adult populations.

## 5. Conclusion

In this research article, we provide a comprehensive overview of key concepts, challenges, and solutions in the domains of AI when applied to pediatric CHD. We emphasize the importance of integrating XAI and FL for enhancing decision support in pediatric echocardiography. The discussion covers the roles of key components, including communication, central authority, local servers, and the user interface, in facilitating data preprocessing, model sharing, and validation, while maintaining model interpretability, privacy, and data security. Clinical use cases showcase the framework's applicability to various tasks, including view recognition, image segmentation, disease state identification, and heart parameter extraction. This research article serves as a valuable resource for researchers and practitioners seeking to develop their own XAI and FL-driven decision support systems in the field of pediatric echocardiography. The complexity of pediatric heart diseases as well as obtaining echocardiography datasets may present challenges to the development of these frameworks. However, it is a significant step toward integrating the algorithms with traditional clinical workflow in a data-centric approach.

## CRediT authorship contribution statement

**Mohamed Yaseen J and Theodor Uden:** Conceptualized and designed the study. **Mohamed Yaseen J and Theodor Uden**: Drafted and wrote the original manuscript. **Mohamed Yaseen J and Steffen Oeltze-Jafra:** Analyzed the technical aspects of the study. **Theodor Uden, Philipp Beerbaum and Thomas Jack:** Contributed to the analysis and investigation of the study's clinical aspects. **Steffen Oeltze-Jafra and Philipp Beerbaum**: Supervised the research, performed investigations, and revised the manuscript. **Steffen Oeltze-Jafra, Philipp Beerbaum and Thomas Jack:** Contributed to the organization of the study, provided critical manuscript revisions and approved the final version for publication. All authors have read and agreed to the final version of the manuscript.

## Declaration of competing interest

The authors declare that they have no known competing financial interests or personal relationships that could have appeared to influence the work reported in this paper.